\documentclass[conference]{IEEEtran}
\IEEEoverridecommandlockouts
\usepackage{cite}
\usepackage{hyperref}
\usepackage{url}
\usepackage{amsmath,amssymb,amsfonts}
\usepackage{algorithmic}
\usepackage{graphicx}
\usepackage{textcomp}
\usepackage{xcolor}
\usepackage{float}
\usepackage{url}
\usepackage{subcaption}
\def\BibTeX{{\rm B\kern-.05em{\sc i\kern-.025em b}\kern-.08em
    T\kern-.1667em\lower.7ex\hbox{E}\kern-.125emX}}
\begin{document}

\title{Benchmarking BERT-based Models for Sentence-level Topic Classification in Nepali Language\\

}

\author{
\IEEEauthorblockN{Nischal Karki, Bipesh Subedi, Prakash Poudyal, Rupak Raj Ghimire, Bal Krishna Bal}
\IEEEauthorblockA{
    Information and Language Processing Research Lab, Kathmandu University\\
    Kavre, Nepal\\
    nischal3158@gmail.com, bipeshrajsubedi@gmail.com, prakash@ku.edu.np, rughimire@gmail.com, bal@ku.edu.np
}
\thanks{Corresponding author: Bal Krishna Bal (bal@ku.edu.np). 
Accepted for presentation at the Regional International Conference on Natural Language Processing (RegICON 2025), Gauhati University, Guwahati, India, November 27--29, 2025.}
}

\maketitle

\begin{abstract}
Transformer-based models such as BERT have significantly advanced Natural Language Processing (NLP) across many languages. However, Nepali, a low-resource language written in Devanagari script, remains relatively underexplored. This study benchmarks multilingual, Indic, Hindi, and Nepali BERT variants to evaluate their effectiveness in Nepali topic classification. Ten pre-trained models, including mBERT, XLM-R, MuRIL, DevBERT, HindiBERT, IndicBERT, and NepBERTa, were fine-tuned and tested on the balanced Nepali dataset containing 25,006 sentences across five conceptual domains and the performance was evaluated using accuracy, weighted precision, recall, F1-score, and AUROC metrics. The results reveal that Indic models, particularly MuRIL-large, achieved the highest F1-score of 90.60\%, outperforming multilingual and monolingual models. NepBERTa also performed competitively with an F1-score of 88.26\%. Overall, these findings establish a robust baseline for future document-level classification and broader Nepali NLP applications.
\end{abstract}


\section{Introduction}
Natural Language Processing (NLP) has progressed rapidly with the emergence of transformer-based pre-trained language models such as Bidirectional Encoder Representations from Transformers (BERT) \cite{devlin2019bert} and its numerous variants. These models have achieved state-of-the-art results across a wide range of NLP tasks, including text classification, question answering, and machine translation. However, much of this progress has been concentrated on high-resource languages, while low-resource languages such as Nepali remain comparatively underexplored.

Nepali, an Indo-Aryan language written in Devanagari script is the official language of Nepal and one of the 22 scheduled languages of India\footnote{\url{https://www.mha.gov.in/sites/default/files/EighthSchedule\_19052017.pdf}}. Despite its cultural and linguistic significance, NLP research in Nepali language poses significant challenges due to the lack of large annotated corpora, limited computational resources, and rich morphological structure\cite{timilsina2022nepberta}. As a result, the development and evaluation of robust Nepali language models remain limited.

The introduction of multilingual pre-trained models such as Multilingual BERT (mBERT) \cite{devlin2019bertpretrainingdeepbidirectional}, XLM-RoBERTa (XLM-R) \cite{conneau2019unsupervised}, and mDeBERTa \cite{he2021debertadecodingenhancedbertdisentangled} has shown potential for cross-lingual transfer learning, where representations learned from high-resource languages can be shared with linguistically related languages. Building on this foundation, region-specific models like Multilingual Representations for Indian Languages (MuRIL) \cite{khanuja2021muril},  HindiBERT, DevBERT \cite{joshi2023l3cubehindbertdevbertpretrainedbert} and IndicBERT \cite{kakwani2020indicnlpsuite} have been developed to better represent Indic languages that share linguistic similarities. Similarly, monolingual models such as NepBERTa \cite{timilsina2022nepberta}, trained exclusively on Nepali text, aim to capture language-specific syntax and semantics more effectively.

Despite such efforts, a systematic comparison of multilingual, Indic, Hindi, and Nepali language models on Nepali text classification tasks remains limited. 
This study addresses this gap by evaluating diverse BERT-based models across multiple language groups on a balanced Nepali corpus spanning five conceptual domains, offering insights into the strengths and limitations of current pre-trained approaches for Nepali language understanding. Furthermore, this study establishes a baseline for future research on document-level classification, where sentence-level insights can be extended to capture broader contextual representations within full articles.

The remainder of this paper is structured as follows. Section \ref{sec:related-works} reviews the related works relevant to this study. Section \ref{sec:methodology} outlines the methodology employed in our research. Section \ref{sec:experimentation} presents the experimental setup and procedures. Section \ref{sec:result-discussion} presents the results and provides a discussion of the findings. Finally, Section \ref{sec:conclusion-future-works} concludes the paper and highlights directions for future research.

\section{Related Works}
\label{sec:related-works}
Text classification in low-resource languages, such as Nepali, has recently received significant attention, and this is based on the need to process richly structured morphological elements, especially in situations with less annotated data. In early approaches, traditional classification algorithms such as Support Vector Machines (SVM) and Naïve Bayes classification were common, with techniques such as TF-IDF or word embedding being applied to extract features. For example, \cite{shahi2018nepali} proposed Nepali text classification based on Naïve Bayes, Neural Network, SVM, which produced encouraging results on small-scale classification, resulting in SVM's better performance in comparison to other models because of its capability to perform with fewer features.

Subsequent studies shifted toward deep learning architectures to better capture sequential dependencies in Nepali text. \cite{basnet2018improving} improved Nepali news recommendation through LSTM-based classification, using word2vec embeddings and outperforming SVM with an accuracy of 84.63\% across eight categories from five popular portals. Similarly, \cite{thapa2021nepali} utilized GloVe embeddings with Long Short-Term Memory (LSTM) networks for classifying 116,736 Nepali news documents into 14 categories, attaining 95.36\% accuracy and surpassing Convolutional Neural Networks (CNN) and Dense Neural Networks (DNN). \cite{wagle2021comparative} extended this by comparing LSTM, Bi-LSTM, and Transformer models on a corpus of around 200k articles across 17 categories, where a pre-trained Transformer fine-tuned after Masked Language Modeling yielded a test F1-score of 95.45\%. \cite{phdthesis} proposed a hybrid LSTM-CNN model, achieving 92.89\% accuracy on nine categories, highlighting the benefits of combining recurrent and convolutional layers for feature extraction in Nepali sequences.

Recently, there has been an inclusion of models based on the transformer architecture, specifically BERT models, in handling the nuances of low-resource Indic languages. \cite{saud2025optimizing} optimized BERT on the Nepali classification task by using stemming techniques and gradient descent optimizers such as AdamW optimizers, which achieved a weighted accuracy of 93.67\% on the classification of news articles. For other related languages, \cite{yadav2025can} proposed maiBERT, which is a BERT model pre-trained on Maithili corpora, that achieved higher accuracy than NepBERTa on the classification of news articles with an accuracy of 87.02\%. For the Marathi language, \cite{mittal2023l3cube} proposed L3Cube-MahaNews, which is a dataset of more than 105k instances of data into 12 classes, testing monolingual models MahaBERT, which performed better on short, medium, and long documents. Additionally, \cite{ahmad2022potrika} constructed Potrika, which is a large Bangla dataset, comprising more than 664,880 articles of eight categories, making it convenient for NLP tasks like classification and summarization.

Despite these contributions, there have been limited works done on comparative evaluations of multilingual, Indic, Hindi, and Nepali-specific BERT-based models on Nepali corpora. Prior works such as \cite{timilsina2022nepberta} compared XLM-R and mBERT without conducting a comprehensive cross-model analysis. Moreover, research on Nepali classification has been less concerned with the comparison of more than a few classes or more than a few models that aren’t transformer models. In this context, our study advances previous works by benchmarking a broad range of BERT-based models, including multilingual, Indic, Hindi, and Nepali-specific variants, on a balanced Nepali corpus spanning five domains. Unlike previous efforts, we further perform sentence-level classification using an annotated and verified dataset, enabling a more granular evaluation of model performance.

\section{Methodology}
\begin{figure*}[htbp]
    \centering
    \includegraphics[width=0.55\textwidth]{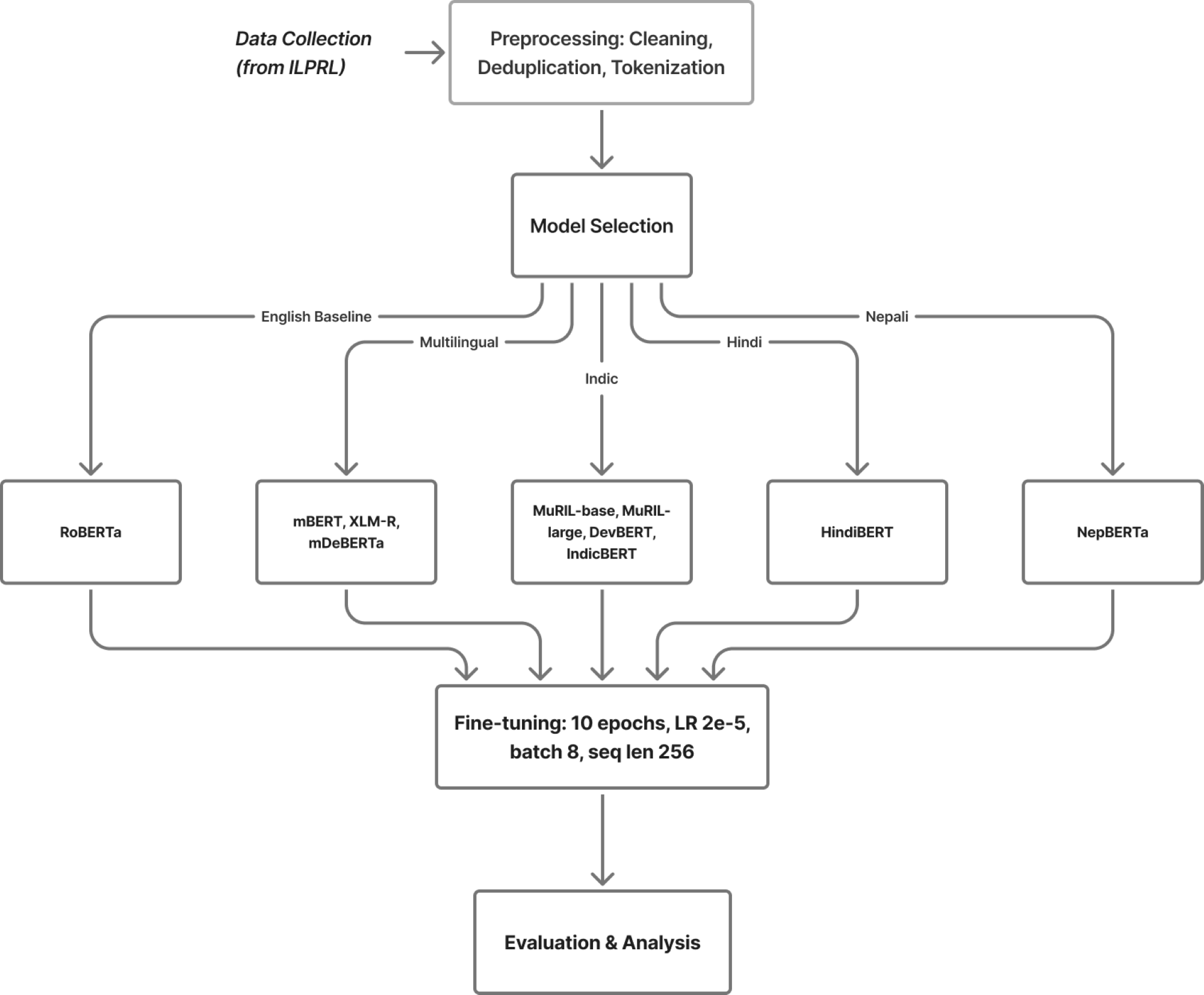} 
    \caption{Overall workflow for Topic Classification}
    \label{fig:methodology}
\end{figure*}

The overall workflow of this study is illustrated in Fig. ~\ref{fig:methodology} which outlines the stages from data collection and preprocessing to model selection, training, and evaluation for topic classification in Nepali text.

\label{sec:methodology}
\subsection{Data Preparation}
The dataset is sourced from the Information and Language Processing Research Lab (ILPRL)\footnote{\url{https://ilprl.ku.edu.np}}
 and consists of around 25K sentences in Nepali language (Devanagari script), distributed across five categories i.e., 1.) Agriculture, 2.) Health, 3.) Education \& Technology 4.) Culture \& Tourism, and 5.) General Communication (derived from Arts, Literature, and Stories). The sentences are filtered and deduplicated to remove any redundancy or noise. The final distribution of the data across categories is presented in Table \ref{tab:data-distribution}.
\begin{table}[htbp]
\centering
\small
\caption{Category-wise distribution of sentences in the dataset.}
\resizebox{\columnwidth}{!}{
\begin{tabular}{|c| c| c| c|}
\hline
\textbf{Code} & \textbf{Category} & \textbf{Sentences} & \textbf{Percentage (\%)} \\
\hline
D1A & Agriculture & 5,012 & 20.04 \\
D2H & Health & 4,999 & 19.99 \\
D3E & Education \& Technology & 4,999 & 19.99 \\
D4C & Culture \& Tourism & 4,998 & 19.99 \\
D5G & General Communication & 4,998 & 19.99 \\
\hline
\textbf{Total} &  & \textbf{25,006} & \textbf{100.0} \\
\hline
\end{tabular}}
\label{tab:data-distribution}
\end{table}

\subsection{Pre-trained Language Models}
In this study, ten BERT-based language models encompassing one English model (\texttt{RoBERTa}), three multilingual models (\texttt{mBERT}, \texttt{XLM-R}, and \texttt{mDeBERTa}), four Indic models (\texttt{MuRIL-base}, \texttt{MuRIL-large}, \texttt{DevBERT}, and \texttt{IndicBERT}), one Hindi model (\texttt{HindiBERT}) and one Nepali model (\texttt{NepBERTa}) are evaluated and compared. The reason behind selecting these models lies in the variation of their pre-training corpora, language coverage, and architectural configurations, offering valuable insights into how multilingual and monolingual representations influence performance in low-resource Nepali text classification task.

\section{Experimentation}
\label{sec:experimentation}
\subsection{Experimental Setup}
For the experiments, all  BERT-based pretrained models were fine-tuned using the Hugging Face Library\footnote{\url{https://huggingface.co/}}. The models were trained on 20,005 sentences, validated on 2,500 sentences, and tested on 2,501 sentences from the dataset. Each model was fine-tuned for \texttt{10 epochs} with a \texttt{learning rate} of \texttt{2e-5}, a \texttt{batch size} of \texttt{8}, \texttt{gradient accumulation} of \texttt{2}, and a \texttt{maximum sequence length} of \texttt{256}.

All of the experiments are performed on a personal desktop equipped with an Intel® Xeon® Platinum 8255C CPU @ 2.50 GHz and a GeForce RTX 4060 Ti 16 GB GPU.

\subsection{Performance Metrics}

Different performance measures are used to provide a comprehensive evaluation of monolingual, multilingual, and Indic models in the topic classification task. Metrics such as accuracy, weighted recall, weighted precision, weighted F1-score, and the area under the receiver operating characteristic curve (AUROC) offer valuable insights into each model’s strengths and weaknesses. 

\section{Result and Discussion}
\label{sec:result-discussion}

Table \ref{tab:weighted_comparison} summarizes the results of our experiments on various pre-trained models, along with their parameter sizes and training-time requirements (in minutes). Among the five groups of BERT, Indic models performed better than other monolingual Nepali,  monolingual English, and  multilingual models. Among Indic models, MuRIL-large outperformed other models across accuracy, weighted precision, weighted recall, weighted F1-score and AUROC metrics with F1-score reaching as high as 90.60\%. In contrast, indicBERT performed the least compared to all other models with weighted F1-score as low as 80.66\%.

\begin{table*}[!ht]
\caption{Performance of Indic, Multilingual, and Nepali Language Models across Different Metrics.}
\label{tab:weighted_comparison}
\centering
\begin{tabular}{|l|c|c|c|c|c|c|c|c|}
\hline

\textbf{Model} & \textbf{\#Params} & \textbf{Training Time (min)} & \textbf{Accuracy (\%)} & \textbf{Precision (\%)} & \textbf{Recall (\%)} & \textbf{AUROC} & \textbf{F1-Score (\%)} \\
\hline
MuRIL (base) & 110M & 49.60 & 89.68 & 89.62 & 89.68 & 0.977 & 89.62 \\
\textbf{MuRIL (large)} &  304M & 65.75 & \textbf{90.56} & \textbf{90.68} & \textbf{90.56} & \textbf{0.984} & \textbf{90.60} \\
DevBERT & 110M & 51.98 & 90.08 & 90.11 & 90.08 & 0.978 & 90.06 \\
IndicBERT & 110M & 29.77 & 80.57 & 80.93 & 80.57 & 0.948 & 80.66 \\
HindiBERT & 110M & 49.45 & 89.68 & 89.65 & 89.68 & 0.975 & 89.63 \\
\hline
mBERT & 177M & 44.68 & 87.84 & 87.75 & 87.84 & 0.970 & 87.78 \\
XLM-R & 278M & 45.13 & 89.72 & 89.67 & 89.72 & 0.979 & 89.67 \\
mDeBERTa & 125M & 39.95 & 86.95 & 86.97 & 86.95 & 0.971 & 86.92 \\
\hline
NepBERTa & 110M & 34.78 & 88.32 & 88.25 & 88.32 & 0.967 & 88.26 \\
\hline
RoBERTa & 125M & 44.28 & 83.81 & 83.81 & 84.09 & 0.966 & 83.83 \\
\hline
\end{tabular}
\end{table*}

\begin{table*}[!ht]
\caption{Performance of Indic, multilingual and Nepali language models across different Categories.}
\label{tab:model_class_comparison}
\centering
\begin{tabular}{|l|c|c|c|c|c|}
\hline
\textbf{Model} & \textbf{F1-score (D1A) (\%)} & \textbf{F1-score (D2H) (\%)} & \textbf{F1-score (D3E) (\%)} & \textbf{F1-score (D4C) (\%)} & \textbf{F1-score (D5G) (\%)} \\
\hline
MuRIL (base) & 92.06 & 87.17 & 91.13 & 94.53 & 82.35 \\
\textbf{MuRIL (large)} & \textbf{93.18} & \textbf{88.87} & \textbf{91.03} & \textbf{94.86} & \textbf{84.37} \\
DevBERT & 92.04 & 87.76 & 91.91 & 93.96 & 83.88 \\
HindiBERT & 92.53 & 87.41 & 90.91 & 94.31 & 82.03 \\
\hline
mBERT & 90.06 & 85.51 & 88.24 & 94.73 & 79.51 \\
XLM-RoBERTa & 91.33 & 87.61 & 90.53 & 94.59 & 83.63 \\
mDeBERTa & 89.81 & 89.41 & 87.14 & 86.37 & 81.69 \\
\hline
NepBERTa & 91.01 & 85.89 & 88.66 & 94.20 & 80.71 \\
\hline
RoBERTa & 84.56 & 80.12 & 85.33 & 93.07 & 75.18 \\
\hline
\end{tabular}
\end{table*}

\begin{figure*}[htbp]
    \centering
    
    \begin{subfigure}{0.23\textwidth}
        \centering
        \includegraphics[width=\linewidth]{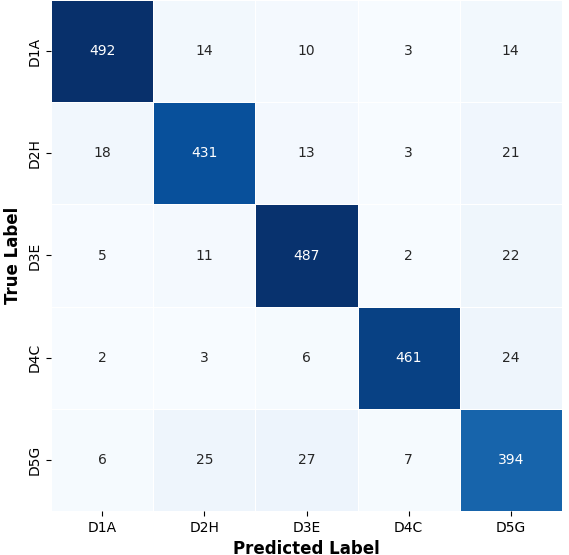}
        \caption{MuRIL-large}
    \end{subfigure}
    \hfill
    \begin{subfigure}{0.23\textwidth}
        \centering
        \includegraphics[width=\linewidth]{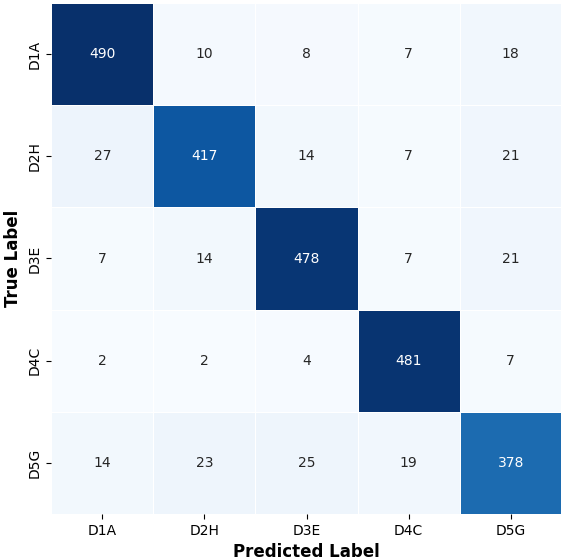}
        \caption{XLM-R}
    \end{subfigure}
    \hfill
    \begin{subfigure}{0.23\textwidth}
        \centering
        \includegraphics[width=\linewidth]{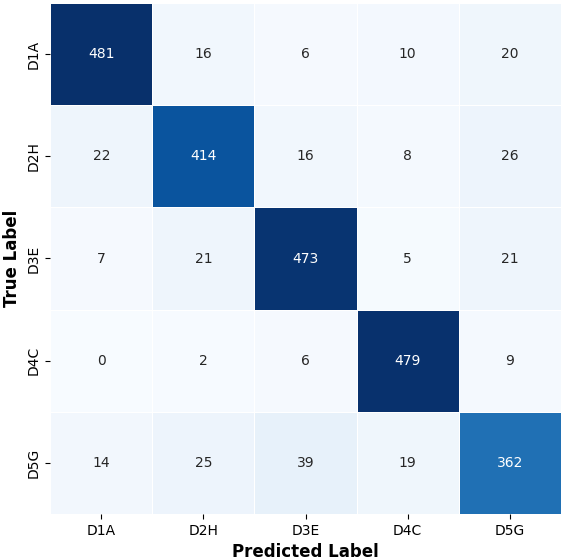}
        \caption{NepBERTa}
    \end{subfigure}
    \hfill
    \begin{subfigure}{0.23\textwidth}
        \centering
        \includegraphics[width=\linewidth]{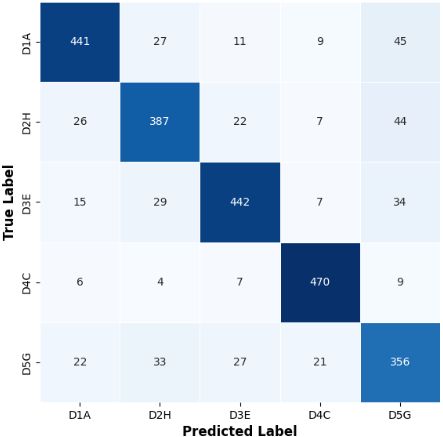}
        \caption{RoBERTa}
    \end{subfigure}

    \caption{Confusion Matrices of the Top-performing Model from Each Group}
    \label{fig:confusion_matrices}
\end{figure*} 

Multilingual models also showed competitive results. XLM-R attained a weighted F1-score of 89.67\%, performing close to the Indic and Hindi-specific models. Meanwhile, mBERT and mdBERTa scored slightly lower, with F1-scores of 87.80\% and 86.92\%, respectively.  However, the English model, RoBERTa lagged behind with an F1-score of 83.83\%.

Furthermore, NepBERTa, trained exclusively on a large Nepali corpus, demonstrated results comparable to the Indic and multilingual models with an F1-score of 88.26\%. Notably, it achieved this at a lower computational cost, requiring only 34.78 minutes of training time despite having just 110M parameters, compared to MuRIL-large’s 304M parameters with 65.75 minutes and XLM-R’s 278M parameters with 45.13 minutes. These results highlight the effectiveness of monolingual pretraining when sufficient Nepali text is available. Nevertheless, although our initial hypothesis was that NepBERTa would outperform Indic and multilingual models due to its Nepali-specific training data, models such as MuRIL-large and XLM-R exhibited slightly stronger results. An explanation to this outcome is that these models were also trained on sizeable amounts of Nepali text \cite{khanuja2021muril, conneau2019unsupervised}.

For category-wise classification, MuRIL-large consistently outperformed the other models, as summarized in Table \ref{tab:model_class_comparison}. The corresponding confusion matrices for the top-performing model in each category are shown in Figure \ref{fig:confusion_matrices}. Taken together, the results indicate that all models are able to classify Agriculture, Education \& Technology, Health, and Culture sentences more accurately compared to General communication. For example, in the confusion matrix for MuRIL-large, 492 of 533 Agriculture-related sentences were correctly classified and 41 were misclassified, whereas only 394 of 459 General Communication-related sentences were correctly classified and 65 were misclassified. The reason behind such a difference may be attributed to the fact that General communication data is derived from multiple sub-categories. However, the results are promising and can be improved in future experiments.

These results reaffirm that models trained on linguistically and scriptually related Indic data (e.g., MuRIL, devBERT, hindBERT) outperform broad multilingual models in representing Nepali text. The consistent advantage of MuRIL-large and NepBERTa supports prior findings that regional and monolingual pretraining captures morphology and syntax more effectively for low-resource languages \cite{khanuja2021muril,timilsina2022nepberta}.

For future researchers, these results highlight the importance of building upon language- and region-specific pretraining. Extending large Indic models like MuRIL through continued pretraining on diverse Nepali corpora or mixed Indic–Nepali datasets could further enhance representational quality. They can also extend  these findings by segmenting documents and applying sentence-level classification to derive document-level insights.

\section*{Limitations}
Although the research provides some valuable insightful information, it has some limitations. The first limitation is that the experiments are confined to text classification and results may not directly generalize to other downstream tasks such as Named-Entity Recognition, Question Answering, or sentiment analysis. The second limitation is the dataset, which although balanced and curated, represents limited domains and text styles, which may restrict broader linguistic variability. Moreover, error analysis among Indic and Nepali models needs to be explored to evaluate models strengths and weaknesses in depth. Lastly, some models like mDeBERTa and DevBERT are still evolving and their training corpora and tokenization schemes are not fully transparent which could influence performance interpretability.

\section{Conclusion and Future Works}
\label{sec:conclusion-future-works}

This study presents the baseline performance of sentence-level news text classification in the Nepali language across five categories by evaluating ten BERT-based models pre-trained on multilingual, Indic, Hindi, and Nepali corpora. The findings illustrate that the Indic models, specifically MuRIL-large, achieved better performance across different metrics despite its larger size and longer training time. NepBERTa offered competitive results with fewer parameters and lower computational cost. The strong performance of MuRIL-large and NepBERTa highlights the effectiveness of region- or language-specific pretraining in low-resource languages like Nepali. Moreover, the findings indicate that our study provides a solid foundation for sentence-level classification, which can later be extended to document-level classification or other applications in the Nepali language.

For the future, we plan to extend the dataset quantitatively with additional categories and conduct detailed error analysis of Indic models and Nepali models. Furthermore, we will investigate how techniques from ensemble learning can be used to increase the classification accuracy. Moreover, investigation on article level classification of different categories remains a possible future direction. This will help us extend our baseline classification problem into a problem of classifying documents, which will give us a richer understanding of the context from which we will be able to generate more accurate topic classification in Nepali texts.

\bibliographystyle{IEEEtran}
\bibliography{reference}

\vspace{12pt}

\end{document}